\UseRawInputEncoding
\documentclass[a4paper,conference]{IEEEtran}

\usepackage{hyperref}
\usepackage{graphicx}
\usepackage{amsmath}
\usepackage{xcolor}

\ifCLASSINFOpdf
\else
\fi

\usepackage{amsmath}

 \usepackage[caption=false,font=footnotesize]{subfig}

\usepackage{dblfloatfix}

\begin{document}
\title{Rethinking CNN Models for Audio Classification}

\author{\IEEEauthorblockN{Kamalesh Palanisamy\IEEEauthorrefmark{1}\IEEEauthorrefmark{2},
Dipika Singhania\IEEEauthorrefmark{2} and
Angela Yao\IEEEauthorrefmark{2}}

\IEEEauthorblockA{\IEEEauthorrefmark{1}
Department of Instrumentation and Control Engineering,\\
National Institute of Technology, Tiruchirappalli, India\\ Email: kamalesh800@gmail.com}
\IEEEauthorblockA{\IEEEauthorrefmark{2} School of Computing\\
National University of Singapore, Singapore\\
Email: \{dipika16, ayao\}@comp.nus.edu.sg }}

\maketitle

\begin{abstract}

In this paper, we show that ImageNet-Pretrained standard deep CNN models can be used as strong baseline networks for audio classification. Even though there is a significant difference between audio Spectrogram and standard ImageNet image samples, transfer learning assumptions still hold firmly. To understand what enables the ImageNet pretrained models to learn useful audio representations, we systematically study how much of pretrained weights is useful for learning spectrograms. We show (1) that for a given standard model using pretrained weights is better than using randomly initialized weights (2) qualitative results of what the CNNs learn from the spectrograms by visualizing the gradients.  Besides, we show that even though we use the pretrained model weights for initialization, there is variance in performance in various output runs of the same model. This variance in performance is due to the random initialization of linear classification layer and random mini-batch orderings in multiple runs.  This brings significant diversity to build stronger ensemble models with an overall improvement in accuracy. An ensemble of ImageNet pretrained DenseNet achieves 92.89\% validation accuracy on the ESC-50 dataset and 87.42\% validation accuracy on the UrbanSound8K dataset which is the current state-of-the-art on both of these datasets.

\end{abstract}

\IEEEpeerreviewmaketitle

\section{Introduction}

To build a model for audio tasks the first step is to decide what kind of representation to use for the data. We can build models using the raw audio waveform \cite{lee2017sample, zhu2016learning} or 2-D representation of the audio like Spectrograms \cite{choi2016automatic, nasrullah2019music, wang2019music}. Spectrograms have become increasingly popular in recent times because they work well with Convolutional Neural Networks(CNN) \cite{choi2016automatic, Dieleman2011AudiobasedMC}. However, CNN models were built for natural images and 2-D spectrograms are different from natural images because natural images contain both space and time information. However, spectrograms contain a temporal dimension it makes them sequential data. Therefore, modifications were suggested to the original CNN architectures. Some created kernels that move along in only one direction to capture temporal data \cite{chen2019cnn}. Other added RNN structure \cite{phan2017audio, gimeno2020multiclass, dai2016long} or Attention \cite{zhang2019attention, wang2019environmental} or a combination of both CNN and RNNs \cite{sang2018convolutional, choi2017convolutional, wang2019music} to improve the sequential understanding of the data. 

In 2014, \cite{gwardys2014deep} showed that we can treat these spectrograms as images and use the standard architecture like AlexNet \cite{krizhevsky2012ImageNet} pretrained on ImageNet \cite{deng2009ImageNet} for audio classification task. The AlexNet model achieved $78\%$ on the GTZAN music genre classification dataset which was the SOTA at the time. However despite there being an improvement in the standard CNN  architectures from AlexNet to ResNet, Inception, DenseNet in the coming years, there has been no work that has used these pretrained ImageNet models for audio tasks.

Most of the works shifted their focus to building models that were more tailored for audio data. Some complicated the entire preprocessing pipeline by using multiple networks to learn different representations of the data \cite{li2019multistream, schindler2018multi} like the raw audio waveform, spectrograms,  MFCCs, etc. The output features from these multiple networks are then aggregated to make the decision. Other papers tried to focus on building custom CNN models \cite{lee2017sample, choi2016automatic, Dieleman2011AudiobasedMC, zhu2016learning} / RNNs \cite{phan2017audio, gimeno2020multiclass, dai2016long} / CRNNs \cite{sang2018convolutional, choi2017convolutional, wang2019music}. Models that were pretrained on large audio datasets like AudioSet\cite{gemmeke2017audio} or the Million Songs Dataset\cite{Bertin-Mahieux2011} were also built. However, people have ignored a strong ImageNet pretrained model baseline to compare the customized models against.

In this paper we show that by using standard architectures like Inception\cite{szegedy2015going}, ResNet\cite{he2016deep}, DenseNet\cite{huang2017densely} pretrained on ImageNet and a single set of input features like Mel-spectrograms we can achieve state-of-the-art results on various datasets like ESC-50 \cite{piczak2015dataset} , UrbanSound8k \cite{Salamon:UrbanSound:ACMMM:14} and above 90\% accuracy on the GTZAN dataset.

The major contributions of this paper are:
\begin{enumerate}
    \item ImageNet pre-trained models fine-tuned for audio datasets can be used to achieve the state of art results and thus can act as a strong baseline that requires minimal feature and model design. We show single hyper-parameter works across all datasets.
    \item We show that various methods used to analyze Transfer Learning \cite{yosinski2014transferable, raghu2019transfusion} for CNNs between different image tasks seem to hold for Transfer Learning between images and spectrograms. 
    \item We use qualitative results based on Integrated Gradients to understand CNN learns the entire shape of the spectrograms. 
\end{enumerate}

\section{Related Work}

\subsection{Audio Classification}

CNN based models have been used for a variety of tasks from Music Genre Classification\cite{dong2018convolutional, choi2016convolutional, Zhang+2016}, Environment Sound Classification\cite{guzhov2020esresnet}\cite{aytar2016soundnet}\cite{9052658} to  Audio Generation\cite{oord2016wavenet, roberts2018hierarchical}. For working with raw audio waveforms, various models that use 1-D convolution have been developed, EnvNet \cite{tokozume2017learning} and Sample-CNN \cite{lee2017sample} are examples of few models that use raw audio as their input. However, most of the SOTA results have been obtained by using CNNs on Spectrograms. Most of these models complicate the design by using multiple models that take different inputs whose outputs are aggregated to make the predictions. For example, \cite{li2019multistream} used three networks to operate on the raw audio, spectrograms, and the delta STFT coefficients;  \cite{su2019environment} used two networks with mel-spectrograms and MFCCs as inputs to the two networks. However, we show that with simple mel-spectrograms one can achieve state-of-the-art performance.

\subsection{Transfer Learning} 

Transfer Learning is the method in which models trained on a particular task with a large amount of data are extended to another task to extract useful features for the new task based on its prior knowledge. In recent years deep models trained on a large corpus like ImageNet for classification have been widely used for transfer learning for tasks such as Image Segmentation \cite{badrinarayanan2017segnet, iglovikov2018ternausnet}, Medical Image Analysis \cite{doi:10.1148/radiol.2019191293, 10.1001/jama.2016.17216}. In video models C3D \cite{carreira2017quo} trained from scratch on UCF-101 \cite{soomro2012ucf101} achieves $88\%$ while pre-training in ImageNet and Kinetics dataset achieves $98\%$ performance. The huge difference in performance between pre-trained weights and training from scratch inspired us to study the difference in Audio Classification. Further, we study details of why ImageNet pre-trained image models are useful for audio classification.

\subsection{Transfer Learning For Audio Classification}

Transfer Learning in Audio-Classification has been mainly focused on pretraining a model on a large corpus of audio datasets like AudioSet, Million Songs Dataset. \cite{choi2017transfer} looked at pre-training a simple CNN network on the Million Song Dataset and found that they can fine-tune these networks for various tasks such as Audio Event Classification, Emotion Prediction; \cite{hershey2016cnn} tried to use large scale models like VGG, Inception \& ResNet for audio classification on AudioSet. However, they trained the models (also called the VGGish) on AudioSet, which is used for many audio transfer learning applications \cite{xie2019zero, shi2019lung}. Different from these, we study transfer learning from massive image datasets like ImageNet.

\subsection{From Image Classification to Audio Classification}

Based on existing work it is clear that transfer learning for audio has focused primarily on audio datasets.  The models used are very large and the features used have also become increasingly complex. As mentioned in the introduction \cite{gwardys2014deep} was one of the first papers to use models pretrained on ImageNet for audio classification.\cite{adapa2019urban, kazakos2019epic, guzhov2020esresnet} has been some of the few works that use models pretrained on ImageNet for audio tasks in recent years. However, these papers did not fully recognize the potential of these models since they made several modifications to the design.  
In this paper, we show that using a single model and a single set of input features we are able to achieve SOTA performance on a variety of tasks thereby reducing the time and space complexity of developing models for audio classification.

\section{Details of Systems and Models}

\begin{figure*}[!t]
\centering
\includegraphics[width=6.75in]{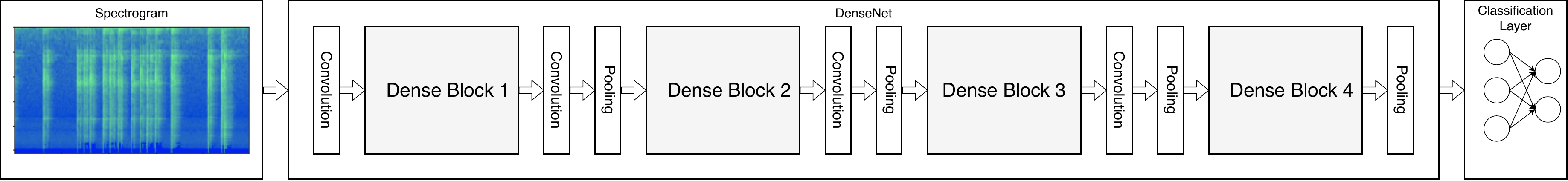}
\caption{\textbf{DenseNet Architecture}: Each Dense Block consists of a certain number of convolution layers whose inputs consist of features from all the previous layers in the block. We use the DenseNet 201 architecture which consists of \{6, 12, 48, 32\} convolution layers in each of the blocks respectively. }
\label{fig_dense}
\end{figure*}

\begin{table*}[!t]
    \centering

    \begin{tabular}{|c|c|c|c|c|c|c|}
         \hline
         Model & \multicolumn{2}{|c|}{GTZAN} & \multicolumn{2}{|c|}{ESC-50} & \multicolumn{2}{|c|}{UrbanSound8K} \\
         \hline 
         & Pretrained & Random & Pretrained & Random & Pretrained & Random \\
         \hline
         DenseNet  & \textbf{91.39\%} & \textbf{88.50\%} & \textbf{91.16\%} & \textbf{72.50\%} & \textbf{85.14\%} & \textbf{76.32\%}\\
         \hline
         ResNet  & 91.09\% & 87.90\% & 90.65\% & 67.40\% & 84.76\% & 73.26\% \\
         \hline
         Inception & 90.00\% & 86.30\% & 87.34\% & 64.50\% & 84.37\% & 75.24\% \\
         \hline
    \end{tabular}
    \caption{Comparison of Accuracy when using Pretrained weights and Random weights. }
    \label{tab:prevsrand}
\end{table*}

\subsection{Datasets}
We tested the models on the following datasets: ESC-50, UrbanSound8K, and the GTZAN dataset. 
\subsubsection{ESC-50}
The Environment Sound Classification(ESC-50)\cite{piczak2015dataset} dataset consists of 2000 clips belonging to 50 classes each of length 5s. The clips are sampled at a uniform rate of 44.1kHz. The dataset is officially split into five-folds, and the accuracy is calculated by cross-validation on all folds. The ESC-50 consists of environmental sounds ranging from sounds of Chirping Birds to Car Horn Sounds. 

\subsubsection{UrbanSound8k}
The UrbanSound8k\cite{Salamon:UrbanSound:ACMMM:14} dataset consists of $8732$ clips belonging to $10$ classes of different urban sounds. Each audio clip is of length $<=4s$, and the sampling rate varies from $16kHz$ to $44.1kHz$. We resampled all audio clips to a sampling rate of $22.5kHz$. The dataset is officially split into $10$ folds, and cross-validation is performed on these $10$ folds.

\subsubsection{GTZAN Dataset}
The GTZAN dataset\footnote{ \hyperlink{http://marsyas.info/downloads/datasets.html} {http://marsyas.info/downloads/datasets.html}} consist of $1000$ music clips each of length $30s$. There are $10$ distinct genre classes. The music clips are sampled at a rate of $22.5kHz$. There is no official training and validation split of the dataset therefore we used $20\%$ of the original data for validation with an equal number of samples for each class and the rest of the data for training.

\subsection{Data Pre-processing}

We performed experiments on the ESC-50 dataset with different representations such as Log-Spectrograms, Log-Melspectrograms, MFCCs, Gammatone-Spectrogram. We used a simple CNN based architecture similar to the 8-layer model of SoundNet\cite{aytar2016soundnet} as a baseline for the experiment. Based on the results which were coherent with \cite{huzaifah2017comparison}, we found out that Log MelSpectrograms were the best feature representation for our particular problem.

CNN based standard models like Resnet, Densenet, Inception use images having three channels as inputs. We need to convert the mel-spectrograms as a three-channel input. We tested two methods in which the input can be given: 
\begin{enumerate}
    \item A single Mel-Spectrogram computed using a window size of $25ms$ and hop length of $10ms$ is replicated across the three channels
    \item The three-channel MelSpectrogram is computed using different window sizes and hop lengths of $\{25ms, 10ms\}$, $\{50ms, 25ms\}$, and $\{100ms, 50ms\}$ on each of the channels respectively. Different window sizes and hop length ensures that the network has different levels of frequency and time information on each channel.
\end{enumerate} 

Based on the baseline model experiments, we find that using mel-spectrograms with different window sizes and hop lengths in each channel gave better performance. These mel-spectrograms were obtained using $128$ mel bins and then log-scaled. Since we used different window sizes, all the mel-spectrograms were reshaped to a common shape. For ESC-50 and UrbanSound8K, we use the input of size $(128, 250)$, whereas, for GTZAN, we use the input of size $(128, 1500)$. 

We use standard Data augmentation techniques such as Time Stretching and Pitch Shifting \cite{Salamon_2017} for the ESC-50 dataset. The data preprocessing was done using Librosa\cite{brian_mcfee_2020_3606573}. 

\subsection{Models}

We used three standard models trained on ImageNet for our problem. The models are:
\begin{enumerate}
    \item Inception\cite{szegedy2015going}: An Inception Layer is a combination of all the layers namely, 1x1 Convolutional layer, 3x3 Convolutional layer, 5x5 Convolutional layers with their output filter banks concatenated into a single output vector forming the input of the next stage. A typical  Inception network consists of several Inception layers stacked upon each other, with occasional max-pooling layers with stride 2 to halve the resolution of the grid. 
    \item ResNet\cite{he2016deep}: ResNet consists of several residual blocks stacked on top of each other. The residual block has two  3x3  convolutional layers with the same number of output channels. Each convolutional layer is followed by a batch normalization layer and a ReLU activation function. A skip connection is added which skips these two convolution operations and adds the input directly before the final ReLU activation function. The objective of the skip connections is to perform identity mapping. 
    \item DenseNet\cite{huang2017densely}: Dense Convolutional Network (DenseNet), connects each layer to every other layer in a feed-forward fashion. For each layer, the feature-maps of all preceding layers are used as inputs, and its own feature-maps are used as inputs into all subsequent layers. Traditional convolutional networks with L layers have L connections — one between each layer and its subsequent layer — a dense network has L(L+1)/ 2 direct connections.

\end{enumerate}

\subsection{Deep Ensemble} We trained $M = \{5\}$ independent models to predict audio classification scores, using the same architecture, hyper-parameter settings, and training procedure as the baseline models. At test time, the ensemble prediction is the average of soft-max outputs of these $M$ individually trained models to evaluate the final accuracy. Independent trained identical models create diversity in ensembles due to differences in model initialization and mini-batch orderings \cite{lakshminarayanan2017simple, kawaguchi2016deep,huang2017snapshot}, which results in different local optimal solutions. We notice here even though we use pre-trained weights for initialization of the convolution network; the linear-classification layer is randomly initialized across various model's runs.

The ensemble model is well known to boost the predictive performance. There are differences in methodologies on how diversity can be added to the ensemble models. \cite{nanni2020ensemble, zahid2015optimized} focuses on adding diversity by using different input samples and different baseline models. Our work differs from these prior works as we focus on the recent finding \cite{lakshminarayanan2017simple, kawaguchi2016deep} that the number of local minimum grows exponentially with the number of parameters used in Neural Network. So without adding any diversity to the modality of the input samples or base model architecture, two identical neural networks, with identical inputs, optimized with different initialization and mini-batch orderings, converge to different solutions. 
\cite{lakshminarayanan2017simple} have shown improvements on standard image datasets, we re-establish its usage in deep models for audio datasets.

\begin{figure*}[!t]
\centering
\subfloat[Weights Change]{\includegraphics[width=2.5in]{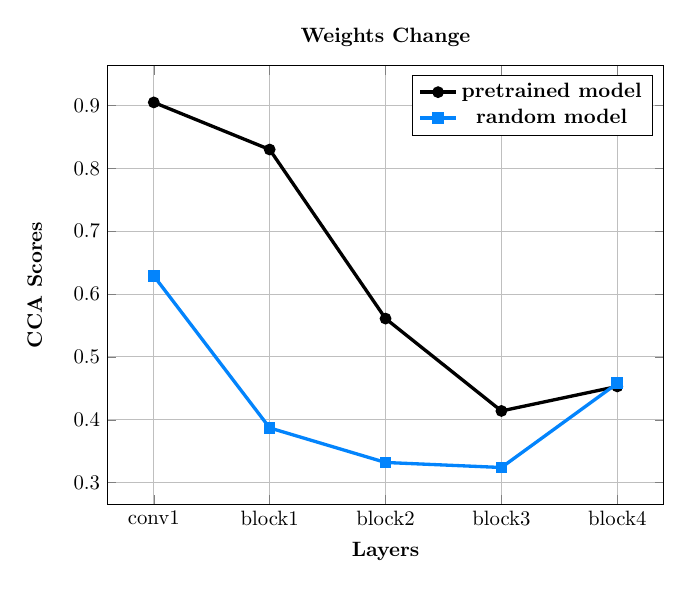}
\label{fig_first_graph}}
\hfil
\subfloat[Weights Fusion]{\includegraphics[width=2.5in]{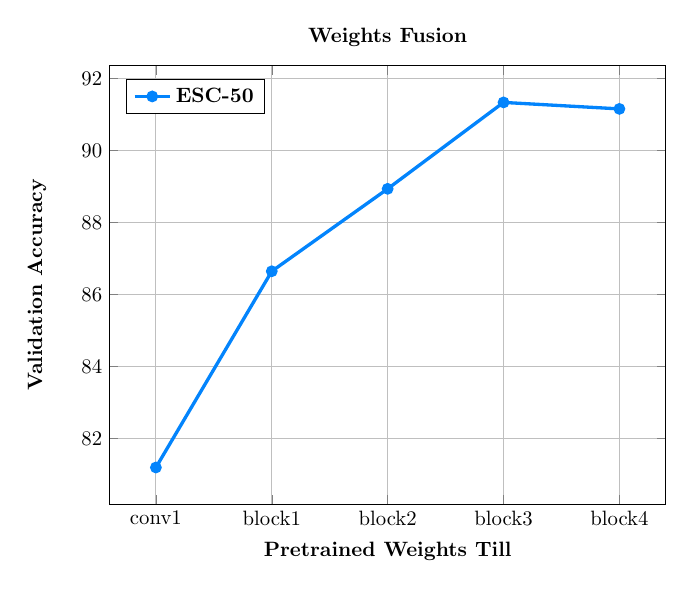}
\label{fig_second_graph}}
\hfil
\\
\subfloat[Weights Freeze]{\includegraphics[width=2.5in]{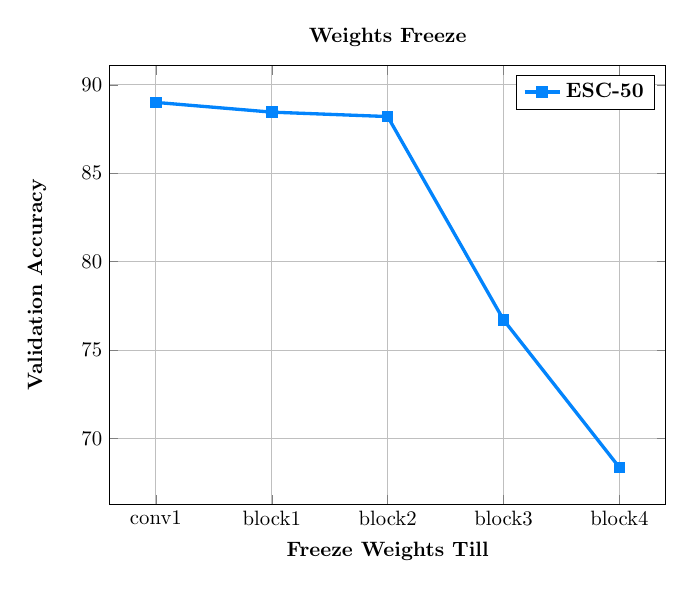}
\label{fig_third_graph}}
\hfil
\caption{\textbf{Analysis of the pretrained models}: (a) CCA similarity scores show that the pretrained models have a high correlation between the weights before and after fine-tuning. For the randomly initialized models, there is a low correlation between the weights before and after fine-tuning (b) The graph shows the validation accuracy for initializing different portions of the network with pretrained weights and initializing the rest of the network with random weights (c) The graph shows the results for freezing a portion of the weights in the pretrained model and fine-tuning the rest of the model}
\label{fig_sim2}
\end{figure*}

\section{Experiments} 

In this section, we will evaluate our models based on experiments conducted. We would evaluate the effectiveness of pre-trained weights, the effectiveness of deep ensemble, and compare our approach to SOTA models.

\subsection{Training the Models}

The optimal hyper-parameter was searched using Grid Search techniques provided by the ray tune library\cite{liaw2018tune}. We found a learning rate of $1e-4$ and a weight decay of $1e-3$ best searched values for training all the models. We used Adam optimizer with a batch size of $32$. All the models were trained using a single Nvidia RTX 2080 GPU.  The code and the checkpoints are available on GitHub\footnote{https://github.com/kamalesh0406/Audio-Classification}.

\subsection{Comparison of fine-tuning ImageNet pre-trained models and Training models from Scratch}

\subsubsection{Setup}

We conducted experiments to understand whether pretrained models are better than randomly initialized models. Each of the ResNet, Inception, and DenseNet models are initialized with pretrained weights and fine-tuned on ESC-50, GTZAN, and UrbanSound8K. The ImageNet pretrained models were trained for $70$ epochs. The learning rate was decreased by a factor of $10$ for every $30$ epoch.

For randomly initialized models ResNet, Inception, and DenseNet are trained from scratch on ESC-50, GTZAN, and UrbanSound8K. In accordance with the training models from scratch for small data regime \cite{yosinski2014transferable}, we trained these models for $450$ epochs and the learning rate decreased by a factor of $10$ at the $300$ and $350$ epoch.  

\subsubsection{Results}

The results of this experiment are shown in Table \ref{tab:prevsrand}. By using pretrained weights we can see a $20\%$ improvement on ESC-50, $10\%$ improvement on UrbanSound8K, and over $3\%$ improvement for the GTZAN dataset. We attribute the difference in results to insufficient data samples available for these small datasets as can been seen in other papers \cite{carreira2017quo}. 

\begin{table*}[!t]
    \centering

    \begin{tabular}{|c|c|c|c|c|c|c|}
         \hline
         Model & \multicolumn{2}{|c|}{GTZAN} & \multicolumn{2}{|c|}{ESC-50} & \multicolumn{2}{|c|}{UrbanSound8K} \\
         \hline 
          & Single & Ensemble & Single & Ensemble & Single & Ensemble\\
         \hline
         DenseNet (Pretrained) & \textbf{91.39$\pm$0.37\%} & 90.50\% & \textbf{91.16$\pm$0.36\%} & \textbf{92.89\%} & \textbf{85.14$\pm$0.17\%} & \textbf{87.42\%}\\
         \hline
         ResNet (Pretrained) & 91.09$\pm$0.86\% & \textbf{91.99\%} & 90.65$\pm$0.28\% & 92.64\% & 84.76$\pm$0.33\% & 87.35\% \\
         \hline
         Inception (Pretrained) & 90.00$\pm$0.70\% & 90.50\% & 87.34$\pm$0.74\% &  89.70\% & 84.37$\pm$0.50\% & 86.34\% \\
         \hline
    \end{tabular}
    \caption{Comparison of Accuracy when using a single model vs ensemble}
    \label{tab:prevsensem}
\end{table*}

\begin{table}[!t]
    \centering

    \begin{tabular}{|c|c|c|c|}
         \hline
         Model & GTZAN & ESC-50 & UrbanSound8K \\
         \hline 
         Choi, Keunwoo, et al.\cite{choi2017transfer} & 89.80\% & - & 69.10\% \\
         \hline
         Multi-Stream Network\cite{li2019multistream} & - & 84.90\% & - \\
         \hline
         Attention-Based CRNN\cite{zhang2019attention} & - & 86.10\% & - \\
         \hline
         ES-ResNet \cite{guzhov2020esresnet} & - & 91.50\% & 85.42\% \\
         \hline
         GTZAN \cite{sturm2013gtzan} & 94.50\% & - & - \\
         \hline
         \hline
         DenseNet (Random) & 88.50\% & 72.50\% & 76.32\%\\
         \hline
         DenseNet (Pretrained) & 91.39\% & 91.16\% & 85.14\%\\
         \hline 
         DenseNet (Pretrained Ensemble) & 90.50\% & \textbf{92.89\%} & \textbf{87.42\%}\\
         \hline
    \end{tabular}
    \caption{Overall Results of Models on three different datasets}
    \label{tab:allresult}
\end{table}
\subsubsection{Analysis of Pre-Trained Weights}

To understand how much of ImageNet pre-trained weights are useful to be transferred to the audio-related task, we conduct the following experiments.
\begin{itemize}
    \item \textit{Setup} All the experiments for transfer learning were performed using the DenseNet Architecture on the ESC-50 dataset. The exact details of how the experiments were conducted are given below:
    \begin{enumerate}
        \item  Weights Change: In the \textbf{Weights Change} experiment, we calculated SVCCA \cite{raghu2017svcca} between the output features of the pretrained network before and after fine-tuning. SVCCA gives a correlation score between the activations of two neurons using Singular Vector Decomposition(SVD). The higher the correlation between the two output features the more similar are the weights of the layers. For our experiments, we use SVCCA to measure the change in weights of the pre-trained network after fine-tuning.
        \item Weight Fusion: In the \textbf{Weight Fusion} experiment, we initialize one portion of the network with pretrained weights and the rest of the network with randomly initialized weight. The entire network is then fine-tuned. 
        \item Weights Freeze: In the \textbf{Weights Freeze} experiment we freeze the weights over a portion of the network and fine-tune the rest of the network. 
        \item Model Cutoff: In the \textbf{Model Cutoff} experiment we remove portions of the network particularly Block4 and Block3 and observe the change in the performance of the network.
        \item Feature Visualization: The visualization experiments involve trying to explain what the network learns from the spectrograms. We use the Integrated Gradients\cite{sundararajan2017axiomatic} method which takes the integral of the gradients of the network with respect to the input and tries to recreate the portions of the input that helps the network make its decision.
    \end{enumerate}
    \item \textit{Results} : The results of the Weights Change experiment is shown in Fig.\ref{fig_first_graph}. The pretrained model shows a high correlation between the features of the initial layers before and after fine-tuning on ESC-50. This suggests that the initial layers of the network undergo little change after fine-tuning. The results of the Weight Fusion experiment is shown in Fig.\ref{fig_second_graph}. We can see that using pretrained weights for up to Block3 has a big impact on the accuracy of the model. The validation accuracy of the model jumps to up to $90\%$ when the Block3 is initialized using the pretrained weights. Beyond Block3, pre-trained weights do not contribute to improvement in results.

    The results of the Weights Change and Weights Fusion experiments suggest that pretrained knowledge is very important in the initial portions of the network. This is because a significant portion of the pretrained knowledge remains in the network suggesting that Spectrograms are treated similarly to Images by the pretrained models. 
    
    The results of the Weights Freeze experiment which is shown in Fig \ref{fig_third_graph} also suggests that Block3 is very important for the network. The accuracy of the network drops by only $2-3\%$ for freezing the first two blocks however it drops by nearly $10\%$ when the weights of Block3 are frozen. Even in the Model Cutoff experiment the accuracy of the network remains to be $90\%$ when the Block4 is removed. However, the validation accuracy drops to be about $85\%$ when we remove both Block3 \& Block4. 
    
    From these experiments we can further pinpoint that Block3 is very important for the network to learn the audio data. These results suggest that the conclusions of \cite{yosinski2014transferable} hold even when we consider transfer learning between two domains with entirely different data. The results of \cite{yosinski2014transferable} suggest that the initial layers of the network contain more general filters and the layers in the middle of the network undergo the most change since they are task-specific. This can be observed in our study where Block 3 seems to be very important for the model to learn features.
    
    The Integrated Gradient visualization for the models is shown in Fig.\ref{fig_sim2}. We can see that networks focus on regions of high energy distribution in the spectrograms. It tries to learn the boundary around these regions similar to how it learns to detect the edges around the objects in the images. Since these boundaries are unique for each sound the network learns to classify them well. Therefore the ImageNet pretrained models which are excellent edge detectors can be easily extended to Spectrograms with sufficient fine-tuning.
\end{itemize}

\begin{figure*}[!t]
\centering
\subfloat{\includegraphics[width=2.5in]{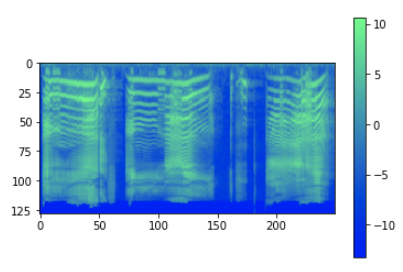}
\label{fig_first_case}}
\hfil
\subfloat{\includegraphics[width=2.5in]{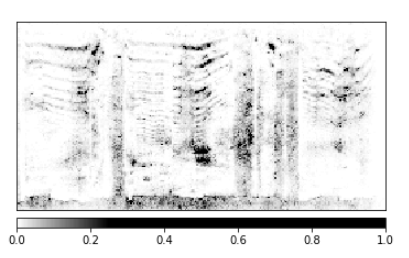}
\label{fig_second_case}}
\hfil
\\
\subfloat{\includegraphics[width=2.5in]{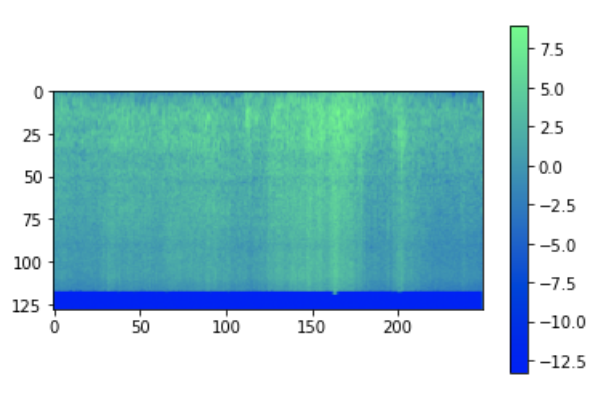}
\label{fig_third_case}}
\hfil
\subfloat{\includegraphics[width=2.5in]{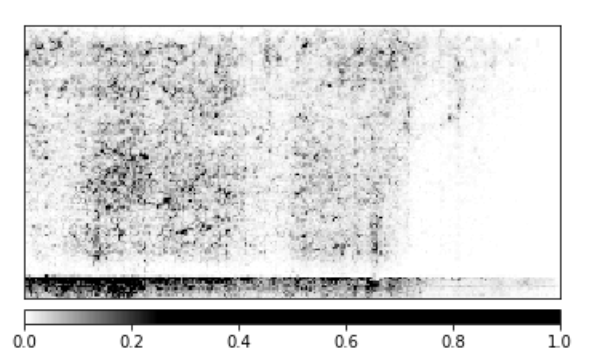}
\label{fig_fourth_case}}
\caption{\textbf{Observing the Integrated Gradients for the Data} The first column shows the data that was given as input to the network and the second column shows the corresponding Integrated Gradients visualization of the input. The Integrated Gradients clearly show us that the model is focusing on the regions where the sound event occurs, this is because the model detects edges around these events and since each of these sounds tends to have a unique shape the model is able to detect them well.}
\label{fig_sim2}
\end{figure*}

\subsection{Deep Ensemble}

\subsubsection{Setup} 

We train 5 independent models with different initialization for the linear layer and different mini-batch orderings. The average of the softmax output of these five models is then taken to produce the ensemble output. The accuracy is calculated using these ensemble outputs. 

\subsubsection{Results}

The results for using ensembling over a single model are shown in Table \ref{tab:prevsensem}. Based on the results we can see that by using ensemble we are able to improve the predictions of the individual models. For both ESC-50 and UrbanSound8k, there is a performance increase of $\sim 2\%$. There is a slight drop in performance for GTZAN as validation data considered for GTZANs consists of only 200 samples, so a drop of $1 \%$ indicates $2$ data samples being incorrectly predicted.

\subsection{Comparison to State-Of-the-Art}

\subsubsection{Comparing Methods} 

On the ESC-50 \& UrbanSound8K the current SOTA model is \cite{guzhov2020esresnet}. \cite{guzhov2020esresnet} built a modification of ResNet and used ImageNet weights to achieve over $91.5\%$ accuracy on the ESC-50 dataset and accuracy of $85.42\%$ on UrbanSound8K dataset. The model they used also contains self-attention layers and for the inputs to their network, they took a spectrogram and split it across its frequency axis and passed it as a three-channel input to the network. For the UrbanSound8K dataset, we have compared our results only with papers that have used the official split provided in the dataset. 

The SOTA accuracy for GTZAN is $94.5\%$, achieved by \cite{sturm2013gtzan}. \cite{sturm2013gtzan} states that a model cannot achieve accuracy greater than $94.5\%$ on the GTZAN dataset due to the noise in the data.

\subsubsection{Results}

The comparison of our models with existing state-of-the-art is shown in Table \ref{tab:allresult}. On the ESC-50 dataset, the ensemble version of DenseNet achieves a validation accuracy of $92.8\%$ and on the UrbanSound8K dataset, the same model achieves a validation accuracy of $87.42\%$ making it the current SOTA models on both the datasets. For the GTZAN dataset, the ensemble version of ResNet can reach an accuracy of  $91.99\%$.

\section{Conclusion}

We proposed that by fine-tuning simple pretrained ImageNet models with a single set of input features for audio tasks we can achieve state-of-the-art results on the ESC-50 and UrbanSound8K dataset and good performance on the GTZAN datasets. We find that the pretrained models retain a major portion of their prior knowledge, especially in the initial layers after fine-tuning. We also find that the intermediate layers of the network undergo significant change to make the model fit the new task. By using qualitative visualizations we demonstrate that the CNN models learn the boundaries of the energy distributions in the spectrograms to classify the spectrograms.

\bibliographystyle{IEEEtran}
\bibliography{References.bib}

\begin{thebibliography}{10}
\providecommand{\url}[1]{#1}
\csname url@samestyle\endcsname
\providecommand{\newblock}{\relax}
\providecommand{\bibinfo}[2]{#2}
\providecommand{\BIBentrySTDinterwordspacing}{\spaceskip=0pt\relax}
\providecommand{\BIBentryALTinterwordstretchfactor}{4}
\providecommand{\BIBentryALTinterwordspacing}{\spaceskip=\fontdimen2\font plus
\BIBentryALTinterwordstretchfactor\fontdimen3\font minus
  \fontdimen4\font\relax}
\providecommand{\BIBforeignlanguage}[2]{{%
\expandafter\ifx\csname l@#1\endcsname\relax
\typeout{** WARNING: IEEEtran.bst: No hyphenation pattern has been}%
\typeout{** loaded for the language `#1'. Using the pattern for}%
\typeout{** the default language instead.}%
\else
\language=\csname l@#1\endcsname
\fi
#2}}
\providecommand{\BIBdecl}{\relax}
\BIBdecl

\bibitem{lee2017sample}
J.~Lee, J.~Park, K.~L. Kim, and J.~Nam, ``Sample-level deep convolutional
  neural networks for music auto-tagging using raw waveforms,'' \emph{arXiv
  preprint arXiv:1703.01789}, 2017.

\bibitem{zhu2016learning}
Z.~Zhu, J.~H. Engel, and A.~Hannun, ``Learning multiscale features directly
  from waveforms,'' \emph{arXiv preprint arXiv:1603.09509}, 2016.

\bibitem{choi2016automatic}
K.~Choi, G.~Fazekas, and M.~Sandler, ``Automatic tagging using deep
  convolutional neural networks,'' \emph{arXiv preprint arXiv:1606.00298},
  2016.

\bibitem{nasrullah2019music}
Z.~Nasrullah and Y.~Zhao, ``Music artist classification with convolutional
  recurrent neural networks,'' in \emph{2019 International Joint Conference on
  Neural Networks (IJCNN)}.\hskip 1em plus 0.5em minus 0.4em\relax IEEE, 2019,
  pp. 1--8.

\bibitem{wang2019music}
Z.~Wang, S.~Muknahallipatna, M.~Fan, A.~Okray, and C.~Lan, ``Music
  classification using an improved crnn with multi-directional spatial
  dependencies in both time and frequency dimensions,'' in \emph{2019
  International Joint Conference on Neural Networks (IJCNN)}.\hskip 1em plus
  0.5em minus 0.4em\relax IEEE, 2019, pp. 1--8.

\bibitem{Dieleman2011AudiobasedMC}
S.~Dieleman, P.~Brakel, and B.~Schrauwen, ``Audio-based music classification
  with a pretrained convolutional network,'' in \emph{ISMIR}, 2011.

\bibitem{chen2019cnn}
M.-T. Chen, B.-J. Li, and T.-S. Chi, ``Cnn based two-stage multi-resolution
  end-to-end model for singing melody extraction,'' in \emph{ICASSP 2019-2019
  IEEE International Conference on Acoustics, Speech and Signal Processing
  (ICASSP)}.\hskip 1em plus 0.5em minus 0.4em\relax IEEE, 2019, pp. 1005--1009.

\bibitem{phan2017audio}
H.~Phan, P.~Koch, F.~Katzberg, M.~Maass, R.~Mazur, and A.~Mertins, ``Audio
  scene classification with deep recurrent neural networks,'' \emph{arXiv
  preprint arXiv:1703.04770}, 2017.

\bibitem{gimeno2020multiclass}
P.~Gimeno, I.~Vi{\~n}als, A.~Ortega, A.~Miguel, and E.~Lleida, ``Multiclass
  audio segmentation based on recurrent neural networks for broadcast domain
  data,'' \emph{EURASIP Journal on Audio, Speech, and Music Processing}, vol.
  2020, no.~1, pp. 1--19, 2020.

\bibitem{dai2016long}
J.~Dai, S.~Liang, W.~Xue, C.~Ni, and W.~Liu, ``Long short-term memory recurrent
  neural network based segment features for music genre classification,'' in
  \emph{2016 10th International Symposium on Chinese Spoken Language Processing
  (ISCSLP)}.\hskip 1em plus 0.5em minus 0.4em\relax IEEE, 2016, pp. 1--5.

\bibitem{zhang2019attention}
Z.~Zhang, S.~Xu, T.~Qiao, S.~Zhang, and S.~Cao, ``Attention based convolutional
  recurrent neural network for environmental sound classification,'' 2019.

\bibitem{wang2019environmental}
H.~Wang, Y.~Zou, D.~Chong, and W.~Wang, ``Environmental sound classification
  with parallel temporal-spectral attention,'' 2019.

\bibitem{sang2018convolutional}
J.~Sang, S.~Park, and J.~Lee, ``Convolutional recurrent neural networks for
  urban sound classification using raw waveforms,'' in \emph{2018 26th European
  Signal Processing Conference (EUSIPCO)}.\hskip 1em plus 0.5em minus
  0.4em\relax IEEE, 2018, pp. 2444--2448.

\bibitem{choi2017convolutional}
K.~Choi, G.~Fazekas, M.~Sandler, and K.~Cho, ``Convolutional recurrent neural
  networks for music classification,'' in \emph{2017 IEEE International
  Conference on Acoustics, Speech and Signal Processing (ICASSP)}.\hskip 1em
  plus 0.5em minus 0.4em\relax IEEE, 2017, pp. 2392--2396.

\bibitem{gwardys2014deep}
G.~Gwardys and D.~M. Grzywczak, ``Deep image features in music information
  retrieval,'' \emph{International Journal of Electronics and
  Telecommunications}, vol.~60, no.~4, pp. 321--326, 2014.

\bibitem{krizhevsky2012ImageNet}
A.~Krizhevsky, I.~Sutskever, and G.~E. Hinton, ``Imagenet classification with
  deep convolutional neural networks,'' in \emph{Advances in neural information
  processing systems}, 2012, pp. 1097--1105.

\bibitem{deng2009ImageNet}
J.~Deng, W.~Dong, R.~Socher, L.-J. Li, K.~Li, and L.~Fei-Fei, ``Imagenet: A
  large-scale hierarchical image database,'' in \emph{2009 IEEE conference on
  computer vision and pattern recognition}.\hskip 1em plus 0.5em minus
  0.4em\relax Ieee, 2009, pp. 248--255.

\bibitem{li2019multistream}
X.~Li, V.~Chebiyyam, and K.~Kirchhoff, ``Multi-stream network with temporal
  attention for environmental sound classification,'' 2019.

\bibitem{schindler2018multi}
A.~Schindler, T.~Lidy, and A.~Rauber, ``Multi-temporal resolution convolutional
  neural networks for acoustic scene classification,'' \emph{arXiv preprint
  arXiv:1811.04419}, 2018.

\bibitem{gemmeke2017audio}
J.~F. Gemmeke, D.~P. Ellis, D.~Freedman, A.~Jansen, W.~Lawrence, R.~C. Moore,
  M.~Plakal, and M.~Ritter, ``Audio set: An ontology and human-labeled dataset
  for audio events,'' in \emph{2017 IEEE International Conference on Acoustics,
  Speech and Signal Processing (ICASSP)}.\hskip 1em plus 0.5em minus
  0.4em\relax IEEE, 2017, pp. 776--780.

\bibitem{Bertin-Mahieux2011}
T.~Bertin-Mahieux, D.~P. Ellis, B.~Whitman, and P.~Lamere, ``The million song
  dataset,'' in \emph{{Proceedings of the 12th International Conference on
  Music Information Retrieval ({ISMIR} 2011)}}, 2011.

\bibitem{szegedy2015going}
C.~Szegedy, W.~Liu, Y.~Jia, P.~Sermanet, S.~Reed, D.~Anguelov, D.~Erhan,
  V.~Vanhoucke, and A.~Rabinovich, ``Going deeper with convolutions,'' in
  \emph{Proceedings of the IEEE conference on computer vision and pattern
  recognition}, 2015, pp. 1--9.

\bibitem{he2016deep}
K.~He, X.~Zhang, S.~Ren, and J.~Sun, ``Deep residual learning for image
  recognition,'' in \emph{Proceedings of the IEEE conference on computer vision
  and pattern recognition}, 2016, pp. 770--778.

\bibitem{huang2017densely}
G.~Huang, Z.~Liu, L.~Van Der~Maaten, and K.~Q. Weinberger, ``Densely connected
  convolutional networks,'' in \emph{Proceedings of the IEEE conference on
  computer vision and pattern recognition}, 2017, pp. 4700--4708.

\bibitem{piczak2015dataset}
\BIBentryALTinterwordspacing
K.~J. Piczak, ``{ESC}: {Dataset} for {Environmental Sound Classification},'' in
  \emph{Proceedings of the 23rd {Annual ACM Conference} on {Multimedia}}.\hskip
  1em plus 0.5em minus 0.4em\relax {ACM Press}, 2015, pp. 1015--1018. [Online].
  Available: \url{http://dl.acm.org/citation.cfm?doid=2733373.2806390}
\BIBentrySTDinterwordspacing

\bibitem{Salamon:UrbanSound:ACMMM:14}
J.~Salamon, C.~Jacoby, and J.~P. Bello, ``A dataset and taxonomy for urban
  sound research,'' in \emph{22nd {ACM} International Conference on Multimedia
  (ACM-MM'14)}, Orlando, FL, USA, Nov. 2014, pp. 1041--1044.

\bibitem{yosinski2014transferable}
J.~Yosinski, J.~Clune, Y.~Bengio, and H.~Lipson, ``How transferable are
  features in deep neural networks?'' in \emph{Advances in neural information
  processing systems}, 2014, pp. 3320--3328.

\bibitem{raghu2019transfusion}
M.~Raghu, C.~Zhang, J.~Kleinberg, and S.~Bengio, ``Transfusion: Understanding
  transfer learning for medical imaging,'' in \emph{Advances in neural
  information processing systems}, 2019, pp. 3347--3357.

\bibitem{dong2018convolutional}
M.~Dong, ``Convolutional neural network achieves human-level accuracy in music
  genre classification,'' 2018.

\bibitem{choi2016convolutional}
K.~Choi, G.~Fazekas, M.~Sandler, and K.~Cho, ``Convolutional recurrent neural
  networks for music classification,'' 2016.

\bibitem{Zhang+2016}
\BIBentryALTinterwordspacing
W.~Zhang, W.~Lei, X.~Xu, and X.~Xing, ``Improved music genre classification
  with convolutional neural networks,'' in \emph{Interspeech 2016}, 2016, pp.
  3304--3308. [Online]. Available:
  \url{http://dx.doi.org/10.21437/Interspeech.2016-1236}
\BIBentrySTDinterwordspacing

\bibitem{guzhov2020esresnet}
A.~Guzhov, F.~Raue, J.~Hees, and A.~Dengel, ``Esresnet: Environmental sound
  classification based on visual domain models,'' 2020.

\bibitem{aytar2016soundnet}
Y.~Aytar, C.~Vondrick, and A.~Torralba, ``Soundnet: Learning sound
  representations from unlabeled video,'' 2016.

\bibitem{9052658}
F.~{Demir}, D.~A. {Abdullah}, and A.~{Sengur}, ``A new deep cnn model for
  environmental sound classification,'' \emph{IEEE Access}, vol.~8, pp.
  66\,529--66\,537, 2020.

\bibitem{oord2016wavenet}
A.~v.~d. Oord, S.~Dieleman, H.~Zen, K.~Simonyan, O.~Vinyals, A.~Graves,
  N.~Kalchbrenner, A.~Senior, and K.~Kavukcuoglu, ``Wavenet: A generative model
  for raw audio,'' \emph{arXiv preprint arXiv:1609.03499}, 2016.

\bibitem{roberts2018hierarchical}
A.~Roberts, J.~Engel, C.~Raffel, C.~Hawthorne, and D.~Eck, ``A hierarchical
  latent vector model for learning long-term structure in music,'' \emph{arXiv
  preprint arXiv:1803.05428}, 2018.

\bibitem{tokozume2017learning}
Y.~Tokozume and T.~Harada, ``Learning environmental sounds with end-to-end
  convolutional neural network,'' in \emph{2017 IEEE International Conference
  on Acoustics, Speech and Signal Processing (ICASSP)}.\hskip 1em plus 0.5em
  minus 0.4em\relax IEEE, 2017, pp. 2721--2725.

\bibitem{su2019environment}
Y.~Su, K.~Zhang, J.~Wang, and K.~Madani, ``Environment sound classification
  using a two-stream cnn based on decision-level fusion,'' \emph{Sensors},
  vol.~19, no.~7, p. 1733, 2019.

\bibitem{badrinarayanan2017segnet}
V.~Badrinarayanan, A.~Kendall, and R.~Cipolla, ``Segnet: A deep convolutional
  encoder-decoder architecture for image segmentation,'' \emph{IEEE
  transactions on pattern analysis and machine intelligence}, vol.~39, no.~12,
  pp. 2481--2495, 2017.

\bibitem{iglovikov2018ternausnet}
V.~Iglovikov and A.~Shvets, ``Ternausnet: U-net with vgg11 encoder pre-trained
  on imagenet for image segmentation,'' \emph{arXiv preprint arXiv:1801.05746},
  2018.

\bibitem{doi:10.1148/radiol.2019191293}
\BIBentryALTinterwordspacing
A.~Majkowska, S.~Mittal, D.~F. Steiner, J.~J. Reicher, S.~M. McKinney, G.~E.
  Duggan, K.~Eswaran, P.-H. Cameron~Chen, Y.~Liu, S.~R. Kalidindi, A.~Ding,
  G.~S. Corrado, D.~Tse, and S.~Shetty, ``Chest radiograph interpretation with
  deep learning models: Assessment with radiologist-adjudicated reference
  standards and population-adjusted evaluation,'' \emph{Radiology}, vol. 294,
  no.~2, pp. 421--431, 2020, pMID: 31793848. [Online]. Available:
  \url{https://doi.org/10.1148/radiol.2019191293}
\BIBentrySTDinterwordspacing

\bibitem{10.1001/jama.2016.17216}
\BIBentryALTinterwordspacing
V.~Gulshan, L.~Peng, M.~Coram, M.~C. Stumpe, D.~Wu, A.~Narayanaswamy,
  S.~Venugopalan, K.~Widner, T.~Madams, J.~Cuadros, R.~Kim, R.~Raman, P.~C.
  Nelson, J.~L. Mega, and D.~R. Webster, ``{Development and Validation of a
  Deep Learning Algorithm for Detection of Diabetic Retinopathy in Retinal
  Fundus Photographs},'' \emph{JAMA}, vol. 316, no.~22, pp. 2402--2410, 12
  2016. [Online]. Available: \url{https://doi.org/10.1001/jama.2016.17216}
\BIBentrySTDinterwordspacing

\bibitem{carreira2017quo}
J.~Carreira and A.~Zisserman, ``Quo vadis, action recognition? a new model and
  the kinetics dataset,'' in \emph{proceedings of the IEEE Conference on
  Computer Vision and Pattern Recognition}, 2017, pp. 6299--6308.

\bibitem{soomro2012ucf101}
K.~Soomro, A.~R. Zamir, and M.~Shah, ``Ucf101: A dataset of 101 human actions
  classes from videos in the wild,'' \emph{arXiv preprint arXiv:1212.0402},
  2012.

\bibitem{choi2017transfer}
K.~Choi, G.~Fazekas, M.~Sandler, and K.~Cho, ``Transfer learning for music
  classification and regression tasks,'' 2017.

\bibitem{hershey2016cnn}
S.~Hershey, S.~Chaudhuri, D.~P.~W. Ellis, J.~F. Gemmeke, A.~Jansen, R.~C.
  Moore, M.~Plakal, D.~Platt, R.~A. Saurous, B.~Seybold, M.~Slaney, R.~J.
  Weiss, and K.~Wilson, ``Cnn architectures for large-scale audio
  classification,'' 2016.

\bibitem{xie2019zero}
H.~Xie and T.~Virtanen, ``Zero-shot audio classification based on class label
  embeddings,'' in \emph{2019 IEEE Workshop on Applications of Signal
  Processing to Audio and Acoustics (WASPAA)}.\hskip 1em plus 0.5em minus
  0.4em\relax IEEE, 2019, pp. 264--267.

\bibitem{shi2019lung}
L.~Shi, K.~Du, C.~Zhang, H.~Ma, and W.~Yan, ``Lung sound recognition algorithm
  based on vggish-bigru,'' \emph{IEEE Access}, vol.~7, pp. 139\,438--139\,449,
  2019.

\bibitem{adapa2019urban}
S.~Adapa, ``Urban sound tagging using convolutional neural networks,'' 2019.

\bibitem{kazakos2019epic}
E.~Kazakos, A.~Nagrani, A.~Zisserman, and D.~Damen, ``Epic-fusion: Audio-visual
  temporal binding for egocentric action recognition,'' in \emph{Proceedings of
  the IEEE International Conference on Computer Vision}, 2019, pp. 5492--5501.

\bibitem{huzaifah2017comparison}
M.~Huzaifah, ``Comparison of time-frequency representations for environmental
  sound classification using convolutional neural networks,'' 2017.

\bibitem{Salamon_2017}
\BIBentryALTinterwordspacing
J.~Salamon and J.~P. Bello, ``Deep convolutional neural networks and data
  augmentation for environmental sound classification,'' \emph{IEEE Signal
  Processing Letters}, vol.~24, no.~3, p. 279–283, Mar 2017. [Online].
  Available: \url{http://dx.doi.org/10.1109/LSP.2017.2657381}
\BIBentrySTDinterwordspacing

\bibitem{brian_mcfee_2020_3606573}
\BIBentryALTinterwordspacing
B.~McFee, V.~Lostanlen, M.~McVicar, A.~Metsai, S.~Balke, C.~Thomé, C.~Raffel,
  A.~Malek, D.~Lee, F.~Zalkow, K.~Lee, O.~Nieto, J.~Mason, D.~Ellis,
  R.~Yamamoto, S.~Seyfarth, E.~Battenberg, Ð.~Морозов, R.~Bittner,
  K.~Choi, J.~Moore, Z.~Wei, S.~Hidaka, nullmightybofo, P.~Friesch, F.-R.
  Stöter, D.~Hereñú, T.~Kim, M.~Vollrath, and A.~Weiss, ``librosa/librosa:
  0.7.2,'' Jan. 2020. [Online]. Available:
  \url{https://doi.org/10.5281/zenodo.3606573}
\BIBentrySTDinterwordspacing

\bibitem{lakshminarayanan2017simple}
B.~Lakshminarayanan, A.~Pritzel, and C.~Blundell, ``Simple and scalable
  predictive uncertainty estimation using deep ensembles,'' in \emph{Advances
  in neural information processing systems}, 2017, pp. 6402--6413.

\bibitem{kawaguchi2016deep}
K.~Kawaguchi, ``Deep learning without poor local minima,'' in \emph{Advances in
  neural information processing systems}, 2016, pp. 586--594.

\bibitem{huang2017snapshot}
G.~Huang, Y.~Li, G.~Pleiss, Z.~Liu, J.~E. Hopcroft, and K.~Q. Weinberger,
  ``Snapshot ensembles: Train 1, get m for free,'' \emph{arXiv preprint
  arXiv:1704.00109}, 2017.

\bibitem{nanni2020ensemble}
L.~Nanni, Y.~M. Costa, R.~L. Aguiar, R.~B. Mangolin, S.~Brahnam, and C.~N.
  Silla, ``Ensemble of convolutional neural networks to improve animal audio
  classification,'' \emph{EURASIP Journal on Audio, Speech, and Music
  Processing}, vol. 2020, no.~1, pp. 1--14, 2020.

\bibitem{zahid2015optimized}
S.~Zahid, F.~Hussain, M.~Rashid, M.~H. Yousaf, and H.~A. Habib, ``Optimized
  audio classification and segmentation algorithm by using ensemble methods,''
  \emph{Mathematical Problems in Engineering}, vol. 2015, 2015.

\bibitem{liaw2018tune}
R.~Liaw, E.~Liang, R.~Nishihara, P.~Moritz, J.~E. Gonzalez, and I.~Stoica,
  ``Tune: A research platform for distributed model selection and training,''
  \emph{arXiv preprint arXiv:1807.05118}, 2018.

\bibitem{sturm2013gtzan}
B.~L. Sturm, ``The gtzan dataset: Its contents, its faults, their effects on
  evaluation, and its future use,'' \emph{arXiv preprint arXiv:1306.1461},
  2013.

\bibitem{raghu2017svcca}
M.~Raghu, J.~Gilmer, J.~Yosinski, and J.~Sohl-Dickstein, ``Svcca: Singular
  vector canonical correlation analysis for deep learning dynamics and
  interpretability,'' 2017.

\bibitem{sundararajan2017axiomatic}
M.~Sundararajan, A.~Taly, and Q.~Yan, ``Axiomatic attribution for deep
  networks,'' 2017.

\end{thebibliography}

\end{document}